\documentclass[final,5p,times,twocolumn,authoryear]{elsarticle}

\usepackage{graphicx}
\usepackage{subcaption}
\usepackage{booktabs}
\usepackage{multirow}
\usepackage{siunitx}
\usepackage{url}
\usepackage{xurl}
\usepackage{hyperref}
\usepackage{xcolor}
\usepackage{colortbl}
\usepackage{amsmath,amssymb}
\usepackage{dblfloatfix}
\graphicspath{
  {./}
  {figs/method_grid_isprs_selected_v3/}
  {figs/method_grid_isprs_selected_v4/stack/}
  {figs/method_grid_isprs_selected_v5/stack/}
  {figs/featcos_v4/}
  {figs/plucker_ray_v1/}
}

\journal{}

\begin{document}

\begin{frontmatter}

\title{EO-VGGT: Orbital Ray-Conditioned 3D Foundation Models for Satellite Multi-View Reconstruction}

\author[label1]{Qiyan Luo}
\author[label1]{Yingdong Pi\corref{cor1}}
\author[label1]{Lekang Wen}
\author[label1]{Jie Yang}
\author[label1]{Xiaoyu Wang}
\author[label1,label2]{Haiming Zhang}
\author[label1,label2]{Mi Wang}

\cortext[cor1]{Corresponding author. Email address: pyd\_imars@whu.edu.cn}

\affiliation[label1]{organization={State Key Laboratory of Information Engineering in Surveying, Mapping and Remote Sensing, Wuhan University},
            city={Wuhan},
            postcode={430079},
            country={China}}
\affiliation[label2]{organization={Hubei Luojia Laboratory},
            city={Wuhan},
            postcode={430079},
            country={China}}

\begin{abstract}

In the era of satellite constellations, multi-view optical satellite imagery is pivotal for Earth Observation (EO) and high-quality Digital Surface Model (DSM) reconstruction. Although feed-forward 3D foundation models have transformed computer vision, their deployment in satellite remote sensing is inherently constrained by the structural discrepancy between implicit perspective assumptions and explicit orbital pushbroom geometry. This geometric incongruity is further compounded by pronounced view-set heterogeneity. We present EO-VGGT, a framework that adapts a frozen perspective-driven model to orbital observations via explicit physical geometry embedding.First, the Geometry-Correlation Constrained Selection (GCCS) strategy prunes sub-optimal observations by balancing geometric diversity and radiometric consistency to optimize the input sequence. Second, a Sensor-Ray Encoder (SRE) parameterizes pixel-level pushbroom lines of sight derived from the Rational Function Model (RFM) into high-dimensional space-geometric tokens, reconciling the mathematical discrepancy between central projection and orbital kinematics. Third, a lightweight Ray-Pointing-Aware Adapter (RPAA) employs gated residual blocks to integrate these tokens directly into the frozen transformer backbone. With under 0.1 \% parameter overhead, this joint conditioning paradigm aligns cross-domain latent representations and facilitates unified forward inference to reconstruct accurate 3D structures.Evaluations on the US3D benchmark demonstrate that EO-VGGT substantially outperforms traditional photogrammetric pipelines, learning-based approaches, and recent neural rendering methods. By overcoming the limited radiometric adaptability and poor cross-domain generalization inherent in these conventional approaches, our framework achieves an 18.0 \% reduction in mean absolute error and an 18.6 \% reduction in the 95th percentile absolute residual $P_{95}\text{Abs}$ compared to the most competitive alternative method while maintaining full dense spatial coverage across heterogeneous radiometric conditions. Our findings underscore that integrating explicit physical geometry with optimized view selection is essential for robust feed-forward satellite 3D reconstruction.

\end{abstract}


\end{frontmatter}

\section{Introduction}

The rapid proliferation of satellite networks and the deployment of dense orbital constellations have initiated a transformative era for modern Earth Observation (EO) \citep{Li03042017,10496159}. Driven by the continuous expansion of these mega-constellations, the unprecedented global coverage and high revisit frequencies provide massive multi-view observation streams. Within this data-rich landscape, leveraging the cross-view synergy of multi-view optical imagery has emerged as a powerful paradigm for high-quality Digital Surface Model (DSM) reconstruction. This capability provides a critical data foundation with substantial potential for large-scale geographic mapping \citep{11196932,8464054,Li_2024_CVPR}, city-scale digital twins \citep{Qian_2023_Sat2Density,Qian_2026_Sat2Densitypp}, and dynamic change monitoring \citep{SemiCDNet,qin20163d}.

Traditionally, satellite multi-view reconstruction relies on a complex, decoupled pipeline involving Structure from Motion (SfM), Bundle Adjustment (BA), and dense matching \citep{8464054,8052151,rs14215617,rs12193158}. Although mature, this sequential paradigm suffers from severe error accumulation. Recent learning-based pipelines and neural rendering paradigms leverage the powerful representation capacity of deep neural networks \citep{zhao_dpsm, stucker_resdepth, huang_mvsr3d, han_satdn,satngp,eogs}. However, these methods frequently suffer from severe performance degradation under orbital environments. Specifically, learning networks often exhibit poor cross-domain generalization across highly heterogeneous geographical regions, while optimization-based neural rendering methods remain highly sensitive to intense radiometric fluctuations and transient shadows characteristic of satellite scenes, thereby limiting their overall robustness \citep{huang2026skysplat}.

Recently, transformer-based feed-forward neural networks have emerged as a highly promising frontier in computer vision. These architectures exhibit an emergent, data-driven capacity for robust geometric reasoning \citep{dust3r,wu_mast3r_eval,vggt,pi3,mapanything}. Models like DUSt3R \citep{dust3r} first revealed the immense potential of large-scale 3D representations. Subsequently, the Visual Geometry Grounded Transformer (VGGT) \citep{vggt} introduced an alternating-attention mechanism to elevate pairwise inference into seamless multi-view 3D scene reconstruction. By bypassing traditional iterative optimizations, these frameworks offer highly generalized forward passes that fundamentally redefine the 3D reconstruction pipeline. Nevertheless, successfully translating this paradigm to large-scale EO mapping remains obstructed by two critical departures from natural-image assumptions \citep{Wu11122025}.

The first departure stems from the breakdown of implicit camera geometry. VGGT inherently relies on implicitly regressing pinhole perspective camera parameters during its forward pass. However, this regression completely fails in orbital configurations. High-resolution optical satellites operate under a line-by-line pushbroom acquisition mechanism, characterized by extreme orbit altitudes and ultra-long focal lengths \citep{gao_rpc_warp}. The resulting imaging geometry is rigorously formulated by the Rational Function Model (RFM) and expressed via Rational Polynomial Coefficients (RPCs). Because this configuration provides no valid perspective priors, the frozen foundation backbone cannot decode orbital camera orientations. This mathematical mismatch completely corrupts the formulation of a coherent global geometric representation. 

The second departure lies in severe view-set heterogeneity. Driven by strict revisit cycles and constellation limitations, satellite stacks exhibit pronounced variations in illumination, multi-date temporal gaps, haze, and cloud occlusions \citep{HSROSS,10641727}. In an end-to-end multi-view network, these low-quality views directly corrupt feature aggregation. The resulting errors propagate throughout the latent space, defying the naive assumption that all input views are uniformly pristine and processed indiscriminately.

Targeting these fundamental theoretical and practical limits, we propose EO-VGGT, a pioneering framework that empowers a frozen perspective-based foundation model to perform reliable satellite multi-view reconstruction. The architecture is structured into a systematic, three-stage decoupled execution pipeline. To resolve the challenge of severe view-set heterogeneity, the pipeline initiates with Pre-Inference View Optimization utilizing a Geometry-Correlation Constrained Selection (GCCS) strategy. Grounded in geometric co-visibility constraints and dense radiometric consistency checking, GCCS prunes degraded observations. This strategy maximizes angular stability across the refined candidate pool to establish an optimal initializer for the global reference frame. Subsequently, to address the breakdown of implicit perspective priors, the framework executes an Orbital Ray Representation stage. Within this stage, a Sensor-Ray Encoder (SRE) explicitly parameterizes pixel-level non-perspective lines of sight from RPC parameters, transforming these physical trajectories into high-dimensional space-geometric tokens. Finally, in the Geometry-Adaptive Surface Reconstruction stage, a lightweight Ray-Pointing-Aware Adapter (RPAA) structurally embeds the derived sensor-ray tokens into the token space of the frozen foundational backbone via gated residuals. By optimizing fewer than 0.1\% of the total network parameters, this explicit geometric injection successfully rectifies the cross-domain mismatch between implicit central-perspective priors and explicit orbital pushbroom observations. This mechanism enables joint forward inference to regress dense 3D point clouds, which are subsequently fused into seamless DSMs.

To summarize, the primary contributions of this paper are organized as follows:
\begin{enumerate}
  \item We propose the Geometry-Correlation Constrained Selection (GCCS) strategy, which filters degraded orbital observations through joint geometric and radiometric constraints to fundamentally suppress latent space error propagation.
  
  \item We design a unified geometric conditioning framework via a Sensor-Ray Encoder (SRE) and a Ray-Pointing-Aware Adapter (RPAA), mapping rigorous satellite sensor rays into high-dimensional tokens to reconcile the mathematical conflict between implicit perspective priors and explicit orbital physics.
  
  \item To our best knowledge, EO-VGGT represents the first endeavor to adapt frozen, general-purpose 3D vision foundation models to satellite multi-view DSM reconstruction, delivering state-of-the-art structural accuracy and continuous spatial completeness across diverse geographical domains.

  \item We develop and evaluate the EO-VGGT framework across a comprehensive benchmark spanning 15 validation scenes and 51 cross-domain testing scenes. The framework achieves an overall Mean Absolute Error of 1.751 meters, representing a 28.5 percent improvement over the unconditioned baseline and an 18.0 percent improvement over the strongest alternative method while outperforming specialized Earth Observation pipelines and maintaining full dense spatial coverage under varying radiometric conditions.
\end{enumerate}

The remainder of this paper is organized as follows. Section~2 reviews related work, and Section~3 details the core EO-VGGT framework. The experimental setup is detailed in Section~4, followed by comprehensive results and analysis in Section~5. Further discussions, limitations, and concluding remarks are distributed across Sections~6 to 8.

\section{Related Work}

\subsection{Feed-Forward 3D Foundation Models}
Driven by large-scale visual priors, feed-forward transformers have recently revolutionized 3D computer vision by bypassing iterative Structure from Motion (SfM) or explicit bundle adjustment algorithms. Leading architectures such as DUSt3R~\citep{dust3r} and MASt3R~\citep{wu_mast3r_eval} directly regress dense 3D coordinate maps and relative camera poses from unconstrained image pairs via cross-view attention mechanics. Similarly, the Visual Geometry Grounded Transformer (VGGT)~\citep{vggt} unifies multi-view matching, track prediction, and camera matrix regression into a seamless, single-forward pass. Subsequent extensions like $\pi^3$~\citep{pi3} and MapAnything~\citep{mapanything} further demonstrate the remarkable data-driven geometric reasoning capabilities of these networks. 

Despite their success, the implicit camera priors of these 3D foundation models are exclusively shaped by perspective, central-projection imagery. When applied directly to orbital remote sensing, these models suffer from catastrophic geometric drift. This failure stems from the fundamental breakdown of the pinhole camera assumption under satellite imaging configurations. Specifically, orbital push-broom sensors utilize linear-array scanning physics that renders the projective rays non-central, which causes perspective focal-length estimations over near-parallel satellite rays to be mathematically ill-conditioned.

\subsection{Satellite Reconstruction Paradigms}
Traditional satellite 3D reconstruction typically operates under a decoupled pipeline comprising epipolar rectification, dense stereo matching, and Digital Surface Model (DSM) fusion. Classical systems instantiate this photogrammetric paradigm through Rational Polynomial Coefficient (RPC)-aware rectification and hand-crafted pairwise matching such as Semi-Global Matching. To alleviate hand-crafted matching errors, subsequent learning-based approaches introduced deep neural networks for confidence-aware stereo or cost-volume regularized multi-view stereo, such as DPSM~\citep{zhao_dpsm}, RESDEPTH~\citep{stucker_resdepth}, MVSR3D~\citep{huang_mvsr3d}, and SatDN~\citep{han_satdn}. Recently, the remote sensing community has witnessed a shift toward coordinate-based neural representations and neural radiance fields, exemplifying frameworks like SAT-NGP~\citep{satngp} and EOGS~\citep{eogs} that unleash fast neural implicit rendering and 3D Gaussian splatting for orbital photogrammetry. 

Although these specialized frameworks handle orbital geometry effectively, they are severely bottlenecked by their heavy-tailed computational overheads, requiring either expensive per-scene optimization or extensive, close-domain training from scratch. This inherently limits their zero-shot generalizability across diverse global geographic terrains. In contrast, our work introduces a new paradigm by efficiently adapting a frozen, pretrained computer vision foundation model such as VGGT to achieve instant, feed-forward satellite reconstruction without any fine-tuning of the core vision backbone.

\subsection{Orbital Geometry Adaptation}
To bridge the mathematical disparities between standard computer vision frameworks and orbital camera models, pioneering efforts focused on adapting perspective-based SfM and MVS to satellite imagery. Frameworks such as SatelliteSfM~\citep{vis_satellite_sfm} achieved this by fitting localized pinhole approximations and perspective proxies directly to RPC metadata, establishing that vision-centric pipelines can deliver competitive reconstruction speed and local morphological accuracy. Nonetheless, these pinhole-fitting approaches are inherently local and highly sensitive to footprint terrain relief, inevitably introducing non-negligible geometric distortions over large Areas of Interest (AOIs). 

To overcome these perspective limitations, subsequent deep-learning frameworks like SatMVS~\citep{gao_rpc_warp} bypassed pinhole approximations entirely by introducing a rigorous RPC warping module. By formulating the Rational Function Model (RFM) as a sequence of tensor transformations, this paradigm successfully integrates native orbital observation geometry directly into learning-based multi-view stereo solvers. Despite their rigorous geometric fidelity, these RPC-warping methods are inextricably bound to traditional dense cost-volume construction, requiring the entire network architecture to be trained from scratch on domain-specific satellite datasets. Consequently, this architecture-dependent warping paradigm cannot be directly adapted to modern frozen 3D foundation models or multi-view transformers without disrupting their emergent pre-trained structural priors or incurring prohibitive retraining overheads. This leaves a critical methodological gap in extending powerful feed-forward foundation models to non-perspective orbital mapping, which demands a more decoupled and parameter-efficient geometry-injection mechanism.

\subsection{Satellite View Selection and Optimization}
Multi-temporal satellite stacks collected across strict revisit cycles inherently suffer from severe acquisition heterogeneity such as pronounced variations in illumination, occlusion, and seasonal land cover. Classical view selection strategies rely primarily on coarse geometric heuristics such as base-to-height ratios to ensure triangulation stability. Early studies on single-pass, in-orbit tri-stereo configurations verified that wider convergence angles bounded height residuals in a manner consistent with traditional aerial photogrammetry~\citep{carl_tristereo,dangelo_safeguards}. However, these frameworks remain constrained to controlled, single-date trajectories and fail to clarify the compounding mechanics under agile cross-track acquisitions. 

Subsequent works established macro-level empirical boundaries for data filtering. For instance, Krau{\ss}~et~al.~\citeyearpar{krauss_cross_track} analyzed the interaction between orbital convergence geometries and illumination discrepancies to identify stable stereo configurations. Crucially, large-scale evaluations by Qin~\citeyearpar{qin_stereo_pairs} demonstrated that solar illumination variance serves as a decisive factor governing radiometric matching consistency and point cloud completeness. Although these classical analyses provide valuable thresholds, they are tailored to local pairwise matching algorithms such as Semi-Global Matching. They fail to model how multi-view radiometric noise disrupts the continuous latent feature aggregation inside end-to-end transformers. To address this limitation, we introduce a Geometry-Correlation Constrained Selection (GCCS) strategy that jointly optimizes geometric diversity and patch-based radiometric correlation. GCCS dynamically prunes low-quality candidates to build a robust multi-view skeleton, preventing error propagation prior to foundation model inference.

\begin{figure*}[t]
  \centering
  \includegraphics[width=\textwidth]{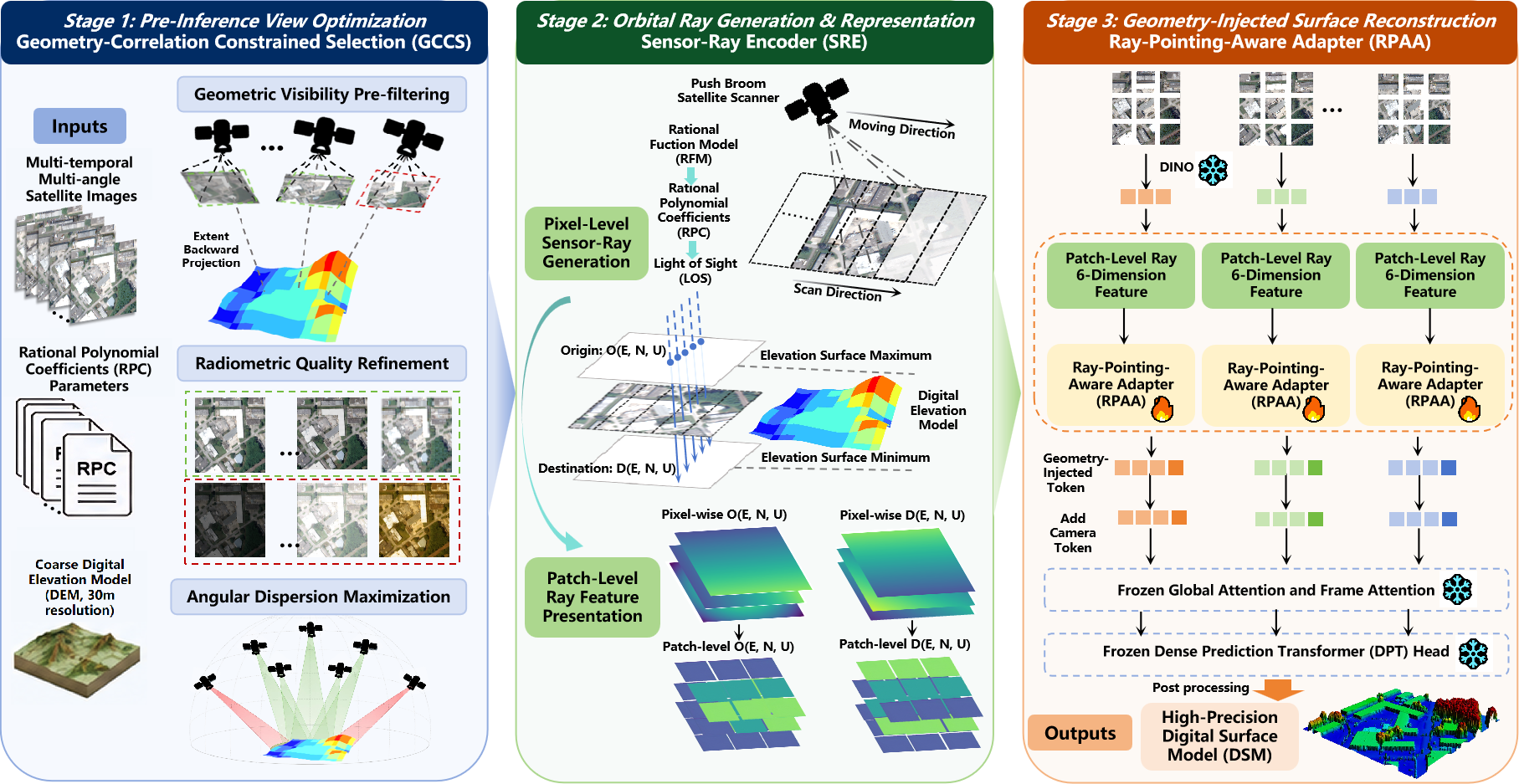}
    \caption{Technical workflow of EO-VGGT for satellite multi-view 3D reconstruction. The framework sequentially filters heterogeneous view stacks via the GCCS strategy, parameterizes rigorous orbital imaging geometry into space-geometric tokens using the Sensor-Ray Encoder (SRE), and structurally injects these physical constraints into the frozen 3D foundation model via the Ray-Pointing-Aware Adapter (RPAA) to regress dense, unbroken DSMs.}
  \label{fig:pipeline}
\end{figure*}

\section{Methodology}

\subsection{Problem Formulation}
Given an Area of Interest (AOI) covered by a collection of $S$ multi-temporal optical satellite views along with their associated RPC camera models, our objective is to reconstruct a dense, high-fidelity Digital Surface Model (DSM) over the geographical footprint. 
In this work, we primarily evaluate the geometric quality of the reconstructed surface after an AOI-level 3D similarity transformation alignment, which isolates the relative structural fidelity from global datum transformations (i.e., scale, rotation, and translation conventions). 
We deliberately focus on relative multi-view reconstruction accuracy rather than absolute geodetic mapping accuracy without prior alignment. While absolute elevation localization is critical for production-ready photogrammetry, it heavily depends on system-level variables such as block adjustment strategies, horizontal geolocation biases, sensor jitters, and ground control points (GCPs)~\citep{SEGPR}. 
Isolating these absolute errors allows for a rigorous and uncorrupted evaluation of the 3D geometric reasoning capabilities of feed-forward foundation models.

\subsection{Methodology Overview}
The overall technical workflow of the proposed EO-VGGT framework is illustrated in Figure~\ref{fig:pipeline}. To adapt a frozen, perspective-driven 3D foundation model to non-perspective orbital remote sensing imagery, the execution pipeline of EO-VGGT is structured into three systematic, decoupled stages:
\begin{enumerate}
    \item Pre-Inference View Optimization. Prior to network execution, the heterogeneous satellite multi-view collection is filtered and curated via the Geometry-Correlation Constrained Selection (GCCS) strategy to establish a radiometrically consistent and geometrically optimal multi-view subset.
    \item Orbital Ray Generation and Representation. To formalize the complex imaging geometry of orbital sensors, pixel-level non-perspective lines of sight are explicitly parameterized based on rational polynomial cameras. These physical lines of sight are subsequently transformed into high-dimensional geometric representations via the Sensor-Ray Encoder (SRE).
    \item Geometry-Injected Surface Reconstruction. The derived sensor-ray representations are structurally embedded into the token space of the frozen foundational backbone via the Ray-Pointing-Aware Adapter (RPAA) to steer cross-view geometric interaction. Conditioned on this explicit geometric injection, the framework executes joint feed-forward inference to regress dense 3D point clouds, which are subsequently rasterized and synthesized into a seamless, area-of-interest-level digital surface model.
\end{enumerate}

\subsection{Geometry-Correlation Constrained Selection (GCCS)}
Multi-temporal satellite images exhibit high acquisition heterogeneity, introducing severe radiometric discrepancies alongside geometric complementarity. Blindly passing uncurated view sequences into multi-view transformers propagates catastrophic latent feature noise. To mitigate this, the Geometry-Correlation Constrained Selection (GCCS) strategy optimizes the input view-set via a decoupled geometric and radiometric pre-filtering pipeline before model execution. For a target tile footprint $\Omega$ with candidate views $\mathcal{V}$ and an anchor ground point $\mathbf{y}_{\Omega}$, GCCS executes a three-stage sequential selection process.

Geometric Visibility Pre-filtering. 
We backproject the local ground footprint bounds into all candidate views using their Rational Polynomial Coefficient (RPC) parameters to construct a geometrically valid candidate pool $\mathcal{V}_{\mathrm{geo}}$:
\begin{equation}
\mathcal{V}_{\mathrm{geo}} = \{v\in\mathcal{V} \mid \pi_v(\mathbf{y}_{\Omega})\in\mathcal{I}_v,\ \rho_v(\Omega)\ge \tau_\rho,\ \theta_v(\mathbf{y}_{\Omega}) \le \tau_\theta\}
\end{equation}
where $\pi_v$ represents the RPC projection into the image domain $\mathcal{I}_v$, $\rho_v(\Omega)$ denotes the sampled footprint coverage, and $\theta_v(\mathbf{y}_{\Omega})$ is the local incidence angle derived from orbital metadata. Views with insufficient coverage or extreme incidence angles are automatically discarded.

Radiometric Quality Refinement. 
To eliminate views corrupted by atmospheric or seasonal variations, we execute a patch-based radiometric consistency check. Taking a near-nadir image as the anchor reference $a$, grayscale patches $P_a$ and $P_v$ of size $p \times p$ are cropped around the projected anchor coordinates for the reference and candidate views, respectively. The Normalized Cross-Correlation (NCC) score $c_v$ is evaluated as:
\begin{equation}
c_v = \frac{\sum_{\mathbf{x}} (P_a(\mathbf{x})-\bar P_a)(P_v(\mathbf{x})-\bar P_v)}
{\sqrt{\sum_{\mathbf{x}}(P_a(\mathbf{x})-\bar P_a)^2}\sqrt{\sum_{\mathbf{x}}(P_v(\mathbf{x})-\bar P_v)^2}}
\end{equation}
where $\bar P_a$ and $\bar P_v$ are spatial patch means. The radiometrically refined pool $\mathcal{V}_{\mathrm{rad}}$ is constructed by retaining the highest-scoring views via $\mathcal{V}_{\mathrm{rad}} = \operatorname{TopM}_{v\in\mathcal{V}_{\mathrm{geo}}}(c_v)$, where $M$ represents the budget of compliant images.

Angular Dispersion Maximization. 
We resolve true space intersection angles $\phi_v = \arccos(\mathbf{d}_a^\top\mathbf{d}_v)$ between candidate sensor lines-of-sight $\mathbf{d}_v$ and the reference direction $\mathbf{d}_a$ via RPC space intersection. An admissible subset $\mathcal{B} = \{v\in\mathcal{V}_{\mathrm{rad}}\setminus\{a\} \mid \tau_{\min}\le\phi_v\le\tau_{\max}\}$ is sorted in ascending order to form an ordered sequence $\mathcal{A} = \operatorname{sort}_{\phi\uparrow}(\mathcal{B})$ of length $m$. To satisfy a strict fixed-budget constraint of $K$ views for joint transformer inference, we uniformly sample the sequence $\mathcal{A}$ using quantile indices $\ell_j$:
\begin{equation}
\ell_j = \operatorname{round}\left(\frac{j(m-1)}{K-2}\right),\quad j=0,\ldots,K-2
\end{equation}
The final optimized input view subset is constructed as $\mathcal{S}_K = \{a\}\cup\{\mathcal{A}_{\ell_j}\}_{j=0}^{K-2}$. This step ensures optimal angular diversity while avoiding low-quality views. All GCCS hyperparameters are fixed a priori to avoid per-AOI test-time tuning.

\subsection{Orbital Ray Generation and Representation}
To formalize the imaging geometry of line-array pushbroom sensors, whose perspective centers continuously evolve along the orbital trajectory, we introduce the Sensor-Ray Encoder (SRE). To avoid the localized geometric distortions inherent in artificial perspective approximations, the SRE maps pixel-level physical imaging rays into high-dimensional geometric tokens by constructing a dense 6D sensor ray map.

For each pixel $(u,v)$ within an image tile, the SRE evaluates the inverse projection of the Rational Function Model across two bounding elevation surfaces, $h_0$ and $h_1$, derived from a coarse local elevation model. The localized metric coordinate origins $\mathbf{o}(u,v)$ and spatial anchors $\mathbf{p}(u,v)$ in a local East-North-Up (ENU) frame are formulated as:
\begin{equation}
\mathbf{o}(u,v) = \mathcal{T}\left(\Pi_{\mathrm{RPC}}^{-1}(u,v,h_0)\right), \quad \mathbf{p}(u,v) = \mathcal{T}\left(\Pi_{\mathrm{RPC}}^{-1}(u,v,h_1)\right)
\end{equation}
where $\Pi_{\mathrm{RPC}}^{-1}$ signifies the RPC backprojection mapping and $\mathcal{T}$ represents the geodetic-to-ENU coordinate transformation. The unit direction vector aligning with the sensor line-of-sight is derived as $\mathbf{d}(u,v) = (\mathbf{p}(u,v)-\mathbf{o}(u,v))/\|\mathbf{p}(u,v)-\mathbf{o}(u,v)\|_2$, yielding the composite 6D sensor ray map $\mathbf{r}(u,v)=[\mathbf{o}(u,v),\mathbf{d}(u,v)]$.

Following channel-wise normalization within the ENU frame, the SRE applies spatial average-pooling with a patch size and stride of $14 \times 14$ pixels to align the continuous 6D ray map with the token resolution of the vision transformer backbone. Within the encoded token space, the first three channels capture the smoothly varying origin trajectories $(o_E,o_N,o_U)$, while the remaining three channels characterize the tokenized line-of-sight direction field $(d_E,d_N,d_U)$. Crucially, these direction channels are treated as learned geometric conditioning descriptors rather than rigid unit vectors, preventing Euclidean norm attenuation during linear aggregation and providing a highly descriptive geometric condition.

\begin{figure}[tbp]
    \centering
    \includegraphics[width=\linewidth]{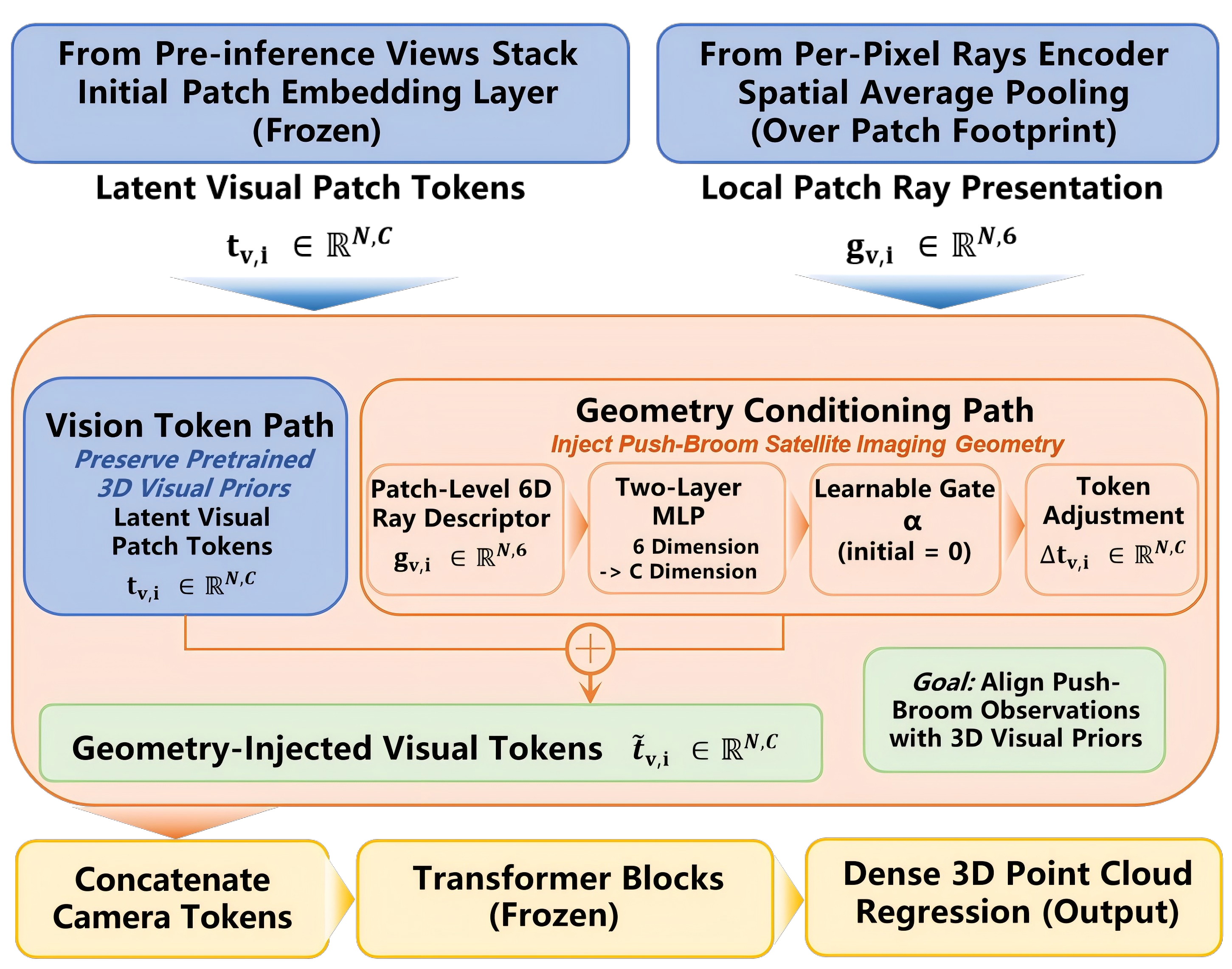}
    \caption{Architecture of the proposed Ray-Pointing-Aware Adapter (RPAA). Patch-level 6D sensor-ray descriptors are projected into the visual token space and fused with visual patch tokens through a gated residual connection before camera-token concatenation and Transformer processing. This lightweight adapter injects orbital imaging geometry into the frozen VGGT backbone while preserving pretrained 3D visual priors.}\label{fig:rpaa}
\end{figure}

\subsection{Geometry-Injected Surface Reconstruction}
To seamlessly incorporate the derived orbital imaging geometry without disrupting the extensive 3D visual priors embedded in the pretrained foundation model, we present the Ray-Pointing-Aware Adapter (RPAA). The detailed architectural layout and data streaming of the RPAA are schematically illustrated in Figure~\ref{fig:rpaa}. The RPAA operates strictly as an input-level patch-token modulation mechanism, ensuring geometric alignment before complex cross-view interactions occur. Let $\mathbf{t}_{v,i}\in\mathbb{R}^{C}$ denote the latent visual patch token extracted by the foundational Visual Geometry Grounded Transformer (VGGT) backbone for view $v$ at patch index $i$. The corresponding localized 6D sensor ray summary pooled over the respective patch footprint is derived via a spatial average pooling layer as: 
\begin{equation}
\mathbf{g}_{v,i}=\mathrm{AvgPool}(\mathbf{r}_v)_i
\end{equation}
where $\mathbf{g}_{v,i} \in \mathbb{R}^{6}$ represents the explicit 6D physical geometric descriptor. 
This descriptor modifies only the raw visual patch tokens immediately following the initial patch embedding layer. Crucially, as indicated in the left panel of Figure~\ref{fig:rpaa}, this modulation is executed prior to the concatenation of camera or register tokens and preceding any global Transformer blocks, thereby preventing the introduction of heavy computational overhead within the internal attention layers.

As the feed-forward data stream enters the core adapter components, the RPAA projects this 6D physical geometric descriptor into the high-dimensional token space through a lightweight two-layer Multi-Layer Perceptron (MLP), generating a structural geometric token adjustment vector: 
\begin{equation}
\Delta\mathbf{t}_{v,i} = \alpha \cdot \mathrm{MLP}(\mathbf{g}_{v,i})
\end{equation}
The geometry-conditioned visual representations are subsequently resolved via a gated residual connection formulated as: 
\begin{equation}
\tilde{\mathbf{t}}_{v,i} = \mathbf{t}_{v,i}+\Delta\mathbf{t}_{v,i}
\end{equation}
where $\alpha \in \mathbb{R}$ is a learnable scalar gating parameter initialized strictly to zero. This zero-initialization strategy ensures that at the onset of training, the framework behaves identically to the vanilla VGGT baseline, preserving structural stability during optimization. Throughout the adaptation phase, the entire pretrained backbone parameters, task-specific attention heads, and prediction interfaces of the foundation model remain frozen. By optimizing solely the minimal parameter set of the RPAA, which constitutes less than 0.1\% of the total parameter budget, the framework successfully rectifies the cross-domain geometric mismatch between central-perspective priors and pushbroom orbital observations.

Conditioned on these geometry-injected tokens ($\tilde{\mathbf{t}}_{v,i}$) processed by the RPAA, the frozen Transformer blocks perform joint cross-view geometric inference to regress dense 3D point clouds. To accommodate expansive geographic regions, the target area of interest is processed via a scalable spatial tiling framework, where point clouds within discrete tiles are rasterized into elevation primitives using intrinsic point-wise confidence metrics as zero-shot uncertainty estimators. Finally, these individual tile-level digital elevation models are seamlessly reconciled and synthesized into a continuous, area-of-interest-level Digital Surface Model (DSM) using a post-inference non-linear quadratic feathered blending scheme, effectively eliminating artificial edge discontinuities and stitching seams.

\subsection{Training Objective}
Following the joint geometric inference, the frozen dense point-regression head decodes the geometry-injected tokens into raw per-pixel 3D world-point predictions. Crucially, because the proposed RPAA solely modulates the input patch-token representations, the framework introduces no auxiliary surface decoders, thereby fully preserving the original deployment topology of the foundation model. To execute structurally consistent supervision free from coordinate system mismatches, a spatial alignment is established between the predicted coordinate space and the ground-truth topography. Specifically, for a randomly sampled subset of $N$ pixels, the corresponding ground-truth elevations are back-projected into 3D metric coordinates using the true RPC models. An optimal 3D similarity transformation, comprising a scale factor $s$, a rotation matrix $\mathbf{R}$, and a translation vector $\mathbf{t}$, is subsequently computed via Umeyama alignment to map the raw predictions into the target frame:
\begin{equation}
\hat{\mathbf{x}}'_j=s\mathbf{R}\hat{\mathbf{x}}_j+\mathbf{t}
\end{equation}
where $\hat{\mathbf{x}}_j$ and $\hat{\mathbf{x}}'_j$ denote the raw and aligned 3D point predictions for pixel $j$, respectively. 

To optimize the adapter parameters, the framework employs a height-focused, alignment-invariant smooth-$L_1$ point loss expressed as:
\begin{equation}
\mathcal{L} = \frac{1}{N}\sum_{j=1}^{N}\rho_{\mathrm{SL1}}(\hat{z}'_j-z_j) +\lambda\frac{1}{N}\sum_{j=1}^{N}\sum_{r\in\{x,y\}}\rho_{\mathrm{SL1}}(\hat{r}'_j-r_j)
\end{equation}
where $\rho_{\mathrm{SL1}}$ denotes the standard smooth-$L_1$ penalty function. The term $z_j$ represents the scalar elevation component, while $(x_j, y_j)$ represents the horizontal coordinate components of the ground-truth 3D point. The balancing hyperparameter is set to $\lambda=0.1$ to bias the optimization primarily toward elevation accuracy while enforcing a soft regularizing constraint on the horizontal coordinates.

\section{Experimental Settings}

\subsection{Dataset and Splits}
We evaluate the performance and cross-domain generalizability of the proposed EO-VGGT framework on the public US3D dataset (DFC2019 Track 3)~\citep{bosch_us3d}. This dataset consists of multi-temporal WorldView-3 satellite image stacks captured over distinct urban areas. To evaluate the framework's robustness against severe radiometric variations, we directly utilize the satellite imagery without applying radiometric calibration or other image processing operations. To rigorously assess model behavior under domain shifts, we implement a strict cross-city evaluation protocol. Specifically, the dataset is partitioned such that Jacksonville (JAX) serves as the source domain for training and validation, whereas Omaha (OMA) is reserved exclusively as an unseen target domain for cross-city generalization testing. Crucially, to eliminate any potential data leakage during model adaptation, the geographic regions within the JAX source domain are divided into strictly spatially non-overlapping tiles or Areas of Interest (AOIs). Consequently, the training set comprises 36 multi-view satellite scenes from JAX, the validation set contains 15 scenes from JAX, and the cross-domain testing set encompasses 51 scenes from OMA. 
The explicit quantitative allocation of the dataset splits is summarized in Table~\ref{tab:splits}.

\begin{table}[t]
  \centering
  \footnotesize
  \caption{Quantitative distribution of the cross-city dataset splits.}
  \label{tab:splits}
  \begin{tabular}{lcc}
    \toprule
    Split Category & Geographic Domain & Number of AOIs \\
    \midrule
    Training & Jacksonville (JAX) & 36 \\
    Validation & Jacksonville (JAX) & 15 \\
    Cross-City Test & Omaha (OMA) & 51 \\
    \bottomrule
  \end{tabular}
\end{table}

\subsection{Compared Methods and Baselines}
We comprehensively benchmark our framework against three distinct groups of contemporary reconstruction pipelines and ablative diagnostic variants:
\begin{enumerate}
    \item General Multi-View Foundation Models. Standard, unconditioned open-source architectures originally trained on perspective central-projection imagery, including VGGT~\citep{vggt}, $\pi^3$~\citep{pi3}, and MapAnything~\citep{mapanything}.
    \item Earth Observation (EO) Specific Pipelines. Traditional photogrammetric and learning-based satellite 3D reconstruction frameworks, including EOGS~\citep{eogs}, SatMVS~\citep{gao_rpc_warp}, and SatNGP~\citep{satngp}. All external pipelines are evaluated using identical geographic coordinates and datum alignment protocols to ensure fair comparison.
    \item Ablative Geometry-Conditioning Diagnostics. Internal variants designed to evaluate the impact of explicit ray structures, specifically comparing our proposed RPC-conditioned adapter framework (EO-VGGT) against a global pinhole-approximated baseline (VGGT+Pinhole).
\end{enumerate}


We categorize the evaluated models into three distinct training and evaluation profiles. First, we evaluate the general multi-view foundation models, which include $\pi^3$, MapAnything, and the vanilla VGGT baseline, under a strict zero-shot setting using their official, unconditioned pretrained weights without any downstream fine-tuning. Second, for our internal diagnostic variants, specifically the proposed EO-VGGT and VGGT+Pinhole, we freeze the multi-billion parameter foundation backbones and exclusively optimize the lightweight structural adapters on the JAX training split. Third, we evaluate the Earth Observation specific references directly using their full-AOI geometric outputs under a unified geographic grid and identical 3D similarity transformation alignments. This multi-tiered benchmarking protocol evaluates our framework against both out-of-the-box foundation generalization and heavily optimized photogrammetric pipelines. For final cross-domain reporting, we select the optimal model parameters determined on the JAX validation set and freeze them for inference across all OMA test scenes.

\subsection{Implementation Details}
The adaptation phase of the proposed Ray-Pointing-Aware Adapter (RPAA) is executed on multi-view spatial patches randomly cropped at a resolution of $512 \times 512$ pixels. During this phase, the pretrained parameters of the core VGGT backbone are entirely frozen, directing the gradient backpropagation exclusively through the lightweight RPAA multi-layer perceptron and its learnable scalar gating parameter $\alpha$. The optimization is conducted using the AdamW optimizer for 800 steps with a learning rate of $3 \times 10^{-4}$ and zero weight decay.

The configuration of the input view stack is governed by the Geometry-Correlation Constrained Selection (GCCS) strategy, which operates in two sequential stages: geometric filtering and radiometric refinement. First, an anchor reference view is selected with a near-nadir preference to maximize geographic co-visibility within each independent tile footprint. Rather than relying on external surveyed ground control points, we sample 512 terrain support points directly from the local DSM grid to serve as a deterministic geometric proxy. These support points are forward-projected into each candidate image via their respective RPC models to estimate the valid footprint coverage ratio $\rho_v(\Omega)$, defined as the fraction of points falling within image boundaries. Views with insufficient tile coverage or an absolute sensor incidence angle $\theta_v$ exceeding $75^\circ$ are systematically pruned. For the remaining candidate pool, a representative tile anchor point is utilized to compute the local line-of-sight direction, absolute incidence angle $\theta_v$, and relative space intersection angle $\phi_v$ relative to the anchor perspective. Crucially, these 512 points function strictly as a pre-inference filter, ensuring zero ground-truth leakage or additional training supervision.

Following the geometric filter, the candidate pool undergoes localized radiometric consistency checking. This stage crops an anchor patch of size $128 \times 128$ pixels, which is subsequently downsampled to $64 \times 64$ pixels to mitigate high-frequency noise during Normalized Cross-Correlation (NCC) filtering. The top $M=18$ radiometrically optimal candidates are preserved in a refinement pool. To bridge this unconditioned pool with downstream token budgets, a quantile dispersion sampler dynamically downsamples the candidates to a final target input view budget of $K=9$. This mechanism sorts the $M$ candidates based on their space intersection angles $\phi_v$ and uniformly subsamples views corresponding to evenly spaced angular quantile indices. The primary angular band for this mapping is configured spanning $[10^\circ, 60^\circ]$, supplemented by an extended fallback range of $[0^\circ, 75^\circ]$. Consequently, the hyperparameter $M$ governs the maximum capacity of the radiometrically trusted observation field, whereas $K$ strictly modulates the geometric density forwarded to the downstream transformer blocks.

\subsection{Evaluation Metrics}
To ensure a fair evaluation across all baselines, all quantitative metrics are computed after spatially aligning the reconstructed 3D models with the ground-truth coordinate system. This alignment protocol eliminates absolute geodetic datum shifts and systematic localization biases, thereby directly capturing the intrinsic structural fidelity of each method. Within this framework, we report a streamlined suite of photogrammetric and geometric metrics. Specifically, the Mean Absolute Error ($\text{MAE}$) quantifies the average vertical displacement across the terrain to provide a linear indicator of overall height deviation, while the Root Mean Squared Error ($\text{RMSE}$) utilizes a quadratic formulation to penalize larger localized errors, identifying severe geometric mismatches or structural artifacts. Furthermore, the 95th percentile of the absolute residual ($P_{95}\text{Abs}$) captures the near-worst-case error bound to reflect reconstruction quality along sharp geometric discontinuities like building facades. Concurrently, the surface completeness ratio ($\text{Completeness}$) measures the percentage of validly reconstructed terrain pixels within the target area. To maintain an unbiased comparison, this $\text{Completeness}$ metric is reported separately from the spatial error metrics ($\text{MAE}$, $\text{RMSE}$, $P_{95}\text{Abs}$). This independent reporting ensures a transparent evaluation for representation frameworks that produce point cloud voids, such as neural rendering or implicit surface extractions, by fully accounting for skipped or highly occluded terrain pixels that escape standard spatial error calculations.

\subsection{Feature Consistency Diagnostic Mechanism}
Beyond macroscopic surface reconstruction errors, we introduce a localized, feature-level diagnostic metric designated as Cross-View Feature Cosine Consistency to evaluate the internal representation alignment within the latent space~\citep{luo2026geometricconsistencyprotocolfoundation}. To directly quantify the internal representation stability across varying orbital observations, this mechanism establishes a geometrically grounded 3D-to-2D projection trace to evaluate whether latent features remain structurally consistent across highly heterogeneous views for the same physical 3D point.

For each sampled reference pixel $\mathbf{x}_i$ in patch coordinates, we obtain its ground-truth height $h_i$ from the Digital Surface Model (DSM) and compute the corresponding 3D space point $\mathbf{X}_i$ via inverse Rational Function Model (RFM) projection in the reference view $r$:
\begin{equation}
    \mathbf{X}_i = \pi^{-1}_{r}(\mathbf{x}_i, h_i)
\end{equation}
In our implementation, this ground point is rigorously solved by a DSM-constrained fixed-point Rational Polynomial Coefficient (RPC) procedure, leveraging the DSM height at the backprojected location as initialization when available. We then forward-project $\mathbf{X}_i$ into each other co-visible target view $j$ using its respective RPC model to obtain the corresponding sub-pixel coordinate $\mathbf{x}_{i}^{(j)} = \pi_{j}(\mathbf{X}_i)$, and bilinearly sample the latent feature map at that exact location.

Let $\mathbf{f}_r(\mathbf{x}_i)$ be the extracted reference feature vector and $\mathbf{f}_j(\mathbf{x}_{i}^{(j)})$ be the matched feature vector in view $j$. Following $L_2$ channel normalization where $\tilde{\mathbf{f}} = \mathbf{f}/\lVert\mathbf{f}\rVert_2$, the per-view cosine similarity is formulated as:
\begin{equation}
    s_{i,j} = \left\langle \tilde{\mathbf{f}}_r(\mathbf{x}_i), \tilde{\mathbf{f}}_j(\mathbf{x}_{i}^{(j)}) \right\rangle
\end{equation}
We then average over the set of valid projected views $\mathcal{V}_i$ to obtain the per-point feature similarity $s_i$:
\begin{equation}
    s_i = \frac{1}{|\mathcal{V}_i|} \sum_{j \in \mathcal{V}_i} s_{i,j}
\end{equation}
To comprehensively quantify the alignment, we report both the overall mean cosine similarity $\frac{1}{N}\sum_{i=1}^N s_i$ across all $N$ valid sampled points, and the class-weighted mean cosine similarity ($\text{clsCos}$) to rigidly account for land-cover semantic variations:
\begin{equation}
    \text{clsCos} = \sum_{c \in \mathcal{C}} \tilde{w}_c \cdot \frac{1}{N_c} \sum_{i:y_i=c} s_i
\end{equation}
where $\mathcal{C} = \{\text{building}, \text{ground}\}$ denotes the designated set of semantic land-cover classes primary focused in this evaluation, $y_i \in \mathcal{C}$ is the semantic label for point $i$, $N_c = |\{i : y_i = c\}|$ is the class-specific point count, and $\tilde{w}_c$ represents the normalized class weights. This diagnostic metric serves as an internal mechanism check to verify whether the explicit injection of orbital imaging rays via the RPAA successfully stabilizes and aligns the latent visual representations across intense radiometric and geometric fluctuations.


\begin{table*}[t]
  \centering
  \caption{Unified overall results under the quadratic feathered blending post-processing protocol. $\pi^3$, MapAnything, and VGGT are general feed-forward foundation-model baselines. All metrics are reported under the macro-averaged AOI mouth at the original ground-truth resolution.}
  \label{tab:main_overall}
  \footnotesize
  \setlength{\tabcolsep}{3.5pt}
  \begin{tabular}{llcccccc}
    \toprule
    Methods & Geometry & Val MAE$\downarrow$ & Test MAE$\downarrow$ & All MAE$\downarrow$ & All RMSE$\downarrow$ & All $P_{95}\text{Abs}$$\downarrow$ & Completeness$\uparrow$ \\
    \midrule
    VGGT & none & 4.039 & 1.980 & 2.448 & 3.957 & 7.921 & 1.0000 \\
    VGGT+pinhole & pinhole rays & 4.021 & 1.974 & 2.439 & 3.951 & 7.907 & 1.0000 \\
    $\pi^3$ & none & 3.858 & 1.630 & 2.136 & 3.913 & 8.202 & 1.0000 \\
    MapAnything & image-only baseline & 3.617 & 2.025 & 2.387 & 4.635 & 8.748 & 0.9980 \\
    EOGS & EO-specific & 5.734 & 7.575 & 7.157 & 9.037 & 17.239 & 0.8810 \\
    SatMVS & EO-specific & 3.052 & 3.288 & 3.234 & 5.054 & 10.064 & 1.0000 \\
    SatNGP & EO-specific & 11.929 & 14.708 & 14.076 & 22.575 & 50.374 & 0.9600 \\
    EO-VGGT (Ours) & RPC rays (RPAA+GCCS) & \textbf{2.547} & \textbf{1.517} & \textbf{1.751} & \textbf{3.184} & \textbf{6.211} & \textbf{1.0000} \\
    \bottomrule
  \end{tabular}
\end{table*}

\begin{table*}[t]
  \centering
  \caption{Full component ablation matrix of RPAA and view selection strategies (evaluated at $K=9$ under the quadratic feathered blending protocol).}
  \label{tab:component_ablation}
  \footnotesize
  \setlength{\tabcolsep}{3.5pt}
  \begin{tabular}{lllccccc}
    \toprule
    Variants & Adapter & View Selection Strategy & Val MAE$\downarrow$ & Test MAE$\downarrow$ & All MAE$\downarrow$ & All RMSE$\downarrow$ & All $P_{95}\text{Abs}$$\downarrow$ \\
    \midrule
    VGGT  & none & random & 3.922 & 2.116 & 2.526 & 3.955 & 7.997  \\
    VGGT & none & base\_angle\_diverse & 4.138 & 1.975 & 2.467 & 3.983 & 7.931  \\
    VGGT & none & GCCS & 4.125 & 1.940 & 2.437 & 3.949 & 7.943  \\
    RPAA & RPAA & random & 2.439 & 1.691 & 1.861 & 3.224 & 6.500  \\
    RPAA & RPAA & base\_angle\_diverse & \textbf{2.534} & 1.533 & 1.761 & 3.184 & 6.234  \\
    EO-VGGT (Ours) & RPAA & GCCS & 2.547 & \textbf{1.517} & \textbf{1.751} & \textbf{3.184} & \textbf{6.211}  \\
    \bottomrule
  \end{tabular}
\end{table*}

\begin{table*}[t]
  \centering
  \caption{Semantic-region results (MAE; macro mean) evaluated using DFC2019 semantic masks under the unified quadratic feathered blending protocol.}
  \label{tab:main_semantic}
  \footnotesize
  \setlength{\tabcolsep}{3.5pt}
  \begin{tabular}{lcccccc}
    \toprule
    Methods & test building$\downarrow$ & test ground$\downarrow$ & test ground+building$\downarrow$ & val building$\downarrow$ & val ground$\downarrow$ & val ground+building$\downarrow$ \\
    \midrule
    VGGT baseline & 1.683 & 2.474 & 1.767 & 3.294 & 5.612 & 4.003 \\
    EO-VGGT pinhole & 1.675 & 2.461 & 1.759 & 3.275 & 5.600 & 3.986 \\
    $\pi^3$ baseline & 1.230 & 2.172 & 1.312 & 3.011 & 5.771 & 3.800 \\
    MapAnything baseline & 1.511 & 2.923 & 1.689 & 2.751 & 5.282 & 3.549 \\
    EOGS rdSM update & 7.481 & 7.685 & 7.439 & 5.464 & 5.692 & 5.575 \\
    SatMVS & 2.439 & 5.601 & 2.844 & 2.018 & 4.935 & 2.913 \\
    SatNGP & 14.526 & 8.782 & 14.338 & 9.789 & 15.808 & 11.131 \\
    EO-VGGT (Ours) & \textbf{1.175} & \textbf{2.005} & \textbf{1.270} & \textbf{1.677} & \textbf{3.977} & \textbf{2.443} \\
    \bottomrule
  \end{tabular}
\end{table*}

\begin{table}[t]
  \centering
  \caption{Cross-view feature cosine consistency for VGGT base and RPC adapter. Random uses natural random sampling; Balanced uses 50\% ground and 50\% building stratified sampling.}
  \label{tab:featcos_overall}
  \footnotesize
  \setlength{\tabcolsep}{3.5pt}
  \begin{tabular}{lcccc}
    \toprule
    \multirow{2}{*}{Split} & \multicolumn{2}{c}{VGGT} & \multicolumn{2}{c}{EO-VGGT} \\
    \cmidrule(lr){2-3} \cmidrule(lr){4-5}
    & Random & Balanced & Random & Balanced \\
    \midrule
    val  & 0.8820 & 0.8837 & 0.9600 & 0.9517 \\
    test & 0.5796 & 0.6146 & 0.8499 & 0.8712 \\
    \bottomrule
  \end{tabular}
\end{table}

\begin{table}[t]
  \centering
  \caption{Region-wise feature cosine consistency under the Balanced protocol.}
  \label{tab:featcos_semantic}
  \footnotesize
  \setlength{\tabcolsep}{3.5pt}
  \begin{tabular}{llcccc}
    \toprule
    Split & Region & Base & Adapter & $\Delta$ & $\Delta\%$ \\
    \midrule
    val  & ground   & 0.8810 & 0.9600 & +0.0790 & +8.97\% \\
    val  & building & 0.8864 & 0.9434 & +0.0570 & +6.43\% \\
    test & ground   & 0.5772 & 0.8479 & +0.2707 & +46.89\% \\
    test & building & 0.6521 & 0.8945 & +0.2424 & +37.18\% \\
    \bottomrule
  \end{tabular}
\end{table}

\begin{table*}[t]
  \centering
  \caption{Detailed view-selection strategy comparison across wide operational budgets ($K$) under the unified quadratic feathered blending protocol. Bold numbers indicate the best performance within each specific budget block ($K$).}
  \label{tab:static_view_selection_k}
  \footnotesize
  \setlength{\tabcolsep}{5pt}
  \begin{tabular}{clccccc}
    \toprule
    $K$ & Strategy & Val MAE$\downarrow$ & Test MAE$\downarrow$ & All MAE$\downarrow$ & All RMSE$\downarrow$ & All $P_{95}\text{Abs}$$\downarrow$ \\
    \midrule
    & random & \textbf{2.463} & 1.645 & 1.830 & \textbf{3.083} & 6.228  \\
    3 & base\_angle\_diverse & 2.535 & \textbf{1.518} & \textbf{1.749} & 3.151 & \textbf{6.161}  \\
      & GCCS & 2.558 & 1.532 & 1.765 & 3.189 & 6.247  \\
    \midrule
    & random & 2.558 & 1.636 & 1.846 & \textbf{3.086} & \textbf{6.078}  \\
    6 & base\_angle\_diverse & 2.556 & \textbf{1.510} & \textbf{1.748} & 3.165 & 6.225  \\
      & GCCS & \textbf{2.550} & 1.523 & 1.756 & 3.195 & 6.224  \\
    \midrule
    & random & \textbf{2.439} & 1.691 & 1.861 & 3.224 & 6.500 \\
    9 & base\_angle\_diverse & 2.534 & 1.533 & 1.761 & \textbf{3.184} & 6.234 \\
      & GCCS (Ours, final configuration) & 2.547 & \textbf{1.517} & \textbf{1.751} & \textbf{3.184} & \textbf{6.211} \\
    \midrule
    & random & \textbf{2.513} & 1.753 & 1.926 & \textbf{3.133} & 6.537 \\
    12 & base\_angle\_diverse & 2.544 & 1.552 & 1.777 & 3.205 & 6.328 \\
       & GCCS & 2.560 & \textbf{1.521} & \textbf{1.757} & 3.198 & \textbf{6.318} \\
    \midrule
    & random & \textbf{2.415} & 1.840 & 1.971 & 3.346 & 6.507 \\
    15 & base\_angle\_diverse & 2.561 & 1.624 & 1.837 & 3.296 & 6.562 \\
       & GCCS & 2.576 & \textbf{1.520} & \textbf{1.760} & \textbf{3.198} & \textbf{6.300} \\
    \midrule
    & random & \textbf{2.522} & 1.943 & 2.074 & 3.475 & 6.926 \\
    18 & base\_angle\_diverse & 2.569 & 1.767 & 1.949 & 3.470 & 6.950 \\
       & GCCS & \textbf{2.556} & \textbf{1.529} & \textbf{1.762} & \textbf{3.184} & \textbf{6.359} \\
    \bottomrule
  \end{tabular}
\end{table*}

\begin{table}[t]
  \centering
  \footnotesize
  \caption{End-to-end computational efficiency profile with and without loading the RPAA adapter. The timing includes complete inference and evaluation over the full 66-AOI benchmark using the same GCCS view-selection cache.}
  \label{tab:runtime_p0}
  \resizebox{\columnwidth}{!}{%
  \begin{tabular}{lccc}
    \toprule
    Profiling metric & w/o RPAA & w/ RPAA & $\Delta$ \\
    \midrule
    Mean time / AOI & 24.4 s & 25.1 s & +0.7 s \\
    Peak GPU memory & 10.42 GiB & 10.52 GiB & +0.10 GiB \\
    \bottomrule
  \end{tabular}%
  }
\end{table}


\section{Results and Analysis}

\subsection{Main Quantitative Benchmarking}
\textbf{Unified Post-Processing and Evaluation Protocol.} We apply a unified post-processing and evaluation protocol to all evaluated models in Table~\ref{tab:main_overall} to ensure a consistent comparison. Each Area of Interest, or AOI, is partitioned into a $3 \times 3$ regular spatial tiling configuration. For general coordinate-free foundation models such as $\pi^3$ and MapAnything, we estimate a rigid 3D similarity transformation matrix using an iterative RANSAC-trimmed Umeyama alignment under a scene-shared spatial reference protocol. This projects the non-metric predictions into the target UTM coordinate system prior to rasterization. All pipelines uniformly export standardized GeoTIFF tiles into an independent merging module that uses a quadratic feathered fusion scheme with a 96-pixel boundary buffer zone to suppress edge discontinuities and block artifacts. We report all primary quantitative metrics using the macro-averaged AOI metric at the original ground-truth resolution across 15 validation scenes and 51 cross-domain testing scenes, resulting in 66 AOIs in total.

\textbf{Overall Performance Analysis.} Table~\ref{tab:main_overall} summarizes the macro-averaged quantitative reconstruction performance under this unified protocol. The proposed EO-VGGT configuration integrates both the Ray-Pointing-Aware Adapter, or RPAA, and the Geometry-Correlation Constrained Selection, or GCCS, strategy with a target view budget of $K=9$. This configuration achieves the highest accuracy across the complete 66-AOI benchmarking pool. It yields an overall Mean Absolute Error, or All MAE, of 1.751 meters, an All RMSE of 3.184 meters, and a 95th percentile absolute residual, denoted as All $P_{95}$, of 6.211 meters. Compared with the unconditioned vanilla VGGT baseline, EO-VGGT reduces the All MAE from 2.448 meters to 1.751 meters, which represents a relative error reduction of 28.5 percent, and lowers the All $P_{95}$ error by 21.6 percent. Compared with external foundation baselines, EO-VGGT lowers the All MAE by 18.0 percent relative to the $\pi^3$ baseline value of 2.136 meters and by 26.6 percent relative to the MapAnything baseline value of 2.387 meters. Furthermore, MapAnything shows an All RMSE of 4.635 meters and an All $P_{95}$ of 8.748 meters, which indicates geometric distortions that our framework avoids.

\textbf{Analysis of the Pushbroom-Perspective Geometric Contradiction.} The performance of the VGGT+Pinhole model yields an All MAE of 2.439 meters, showing a minor improvement of 0.009 meters over the unconditioned vanilla baseline. This indicates a geometric mismatch between the implicit pinhole perspective prior in computer vision architectures and the line-by-line pushbroom acquisition mechanism of orbital imaging sensors. Conventional pinhole approximations assume a single static projection center where all imaging rays converge. In contrast, high-resolution optical satellite arrays capture imagery continuously along an orbital trajectory, meaning that each image row corresponds to a distinct instantaneous optical center and time step. Modeling this moving imaging line with a static pinhole framework introduces non-linear epipolar distortions and systematic spatial warps. Our framework uses the Sensor-Ray Encoder, or SRE, to parameterize Rational Function Model, or RFM, coordinates into high-dimensional space-geometric tokens. This physical conditioning ensures that the latent tokens reflect the true orbital kinematics. Consequently, EO-VGGT achieves a lower error rate by combining the SRE and RPAA modules.

\textbf{Comparison with Earth Observation Baselines.} We evaluate the Earth Observation specific baselines, which include EOGS, SatMVS, and SatNGP, without scene-specific retraining to test their direct operational capability. Conventional handcrafted photogrammetric pipelines often yield sparse reconstruction topologies. These traditional methods lower their error metrics by omitting low-confidence correspondences in textureless regions, deep building shadows, or areas with severe multi-temporal radiometric variations, which restricts the error computation to high-confidence salient features. EO-VGGT achieves an All MAE of 1.751 meters while maintaining complete dense spatial coverage across all regions. Neural rendering methods such as EOGS and SatNGP optimize their representations on a per-scene basis and can generate accurate geometries under strictly stable radiometric conditions. However, multi-view satellite scenes captured at different times inherently lack radiometric consistency due to changing atmospheric, seasonal, and illumination conditions. Because these neural rendering algorithms rely on the strict assumption of photometric consistency across views, they fail to resolve these inherent multi-temporal radiometric discrepancies. This algorithmic limitation leads to severe artifacts, geometric distortions, and warped terrain structures in their reconstructions. While SatMVS enforces spatial completeness, it yields an elevated MAE of 3.234 meters, and SatNGP yields an MAE of 14.076 meters due to these unhandled radiometric variations. EO-VGGT combines full dense spatial coverage with structural accuracy under changing radiometric conditions.

\subsection{Qualitative and Visual Comparison}
Figure~\ref{fig:qual_all} displays the reconstructed DSM geometries across different evaluation scenes. The unconditioned computer vision baselines exhibit large-scale regional tilting, shearing, and structural distortions because they rely on incorrect central-perspective assumptions. EO-VGGT avoids these regional warps, reconstructing flat, stable, and sharply bounded urban topographies. In the close-up regions, our framework recovers accurate building footprints and precise vertical roof facets, preserving sharp geometric boundaries where general foundation architectures produce blurred or dissolved structures. This consistent visual quality across both the JAX source domain and the unseen OMA target domain demonstrates the robust geographic generalization of our explicit ray-conditioning framework. Furthermore, our geometry-correlation constrained selection strategy actively suppresses spatial inconsistencies during the inference process. By ensuring rigorous geometric alignment across multiple views at the algorithmic level, our framework generates high-precision and highly reliable DSMs.

\begingroup
\setlength{\dbltextfloatsep}{3pt}
\begin{figure*}[!ht]
  \centering
  \includegraphics[width=\textwidth]{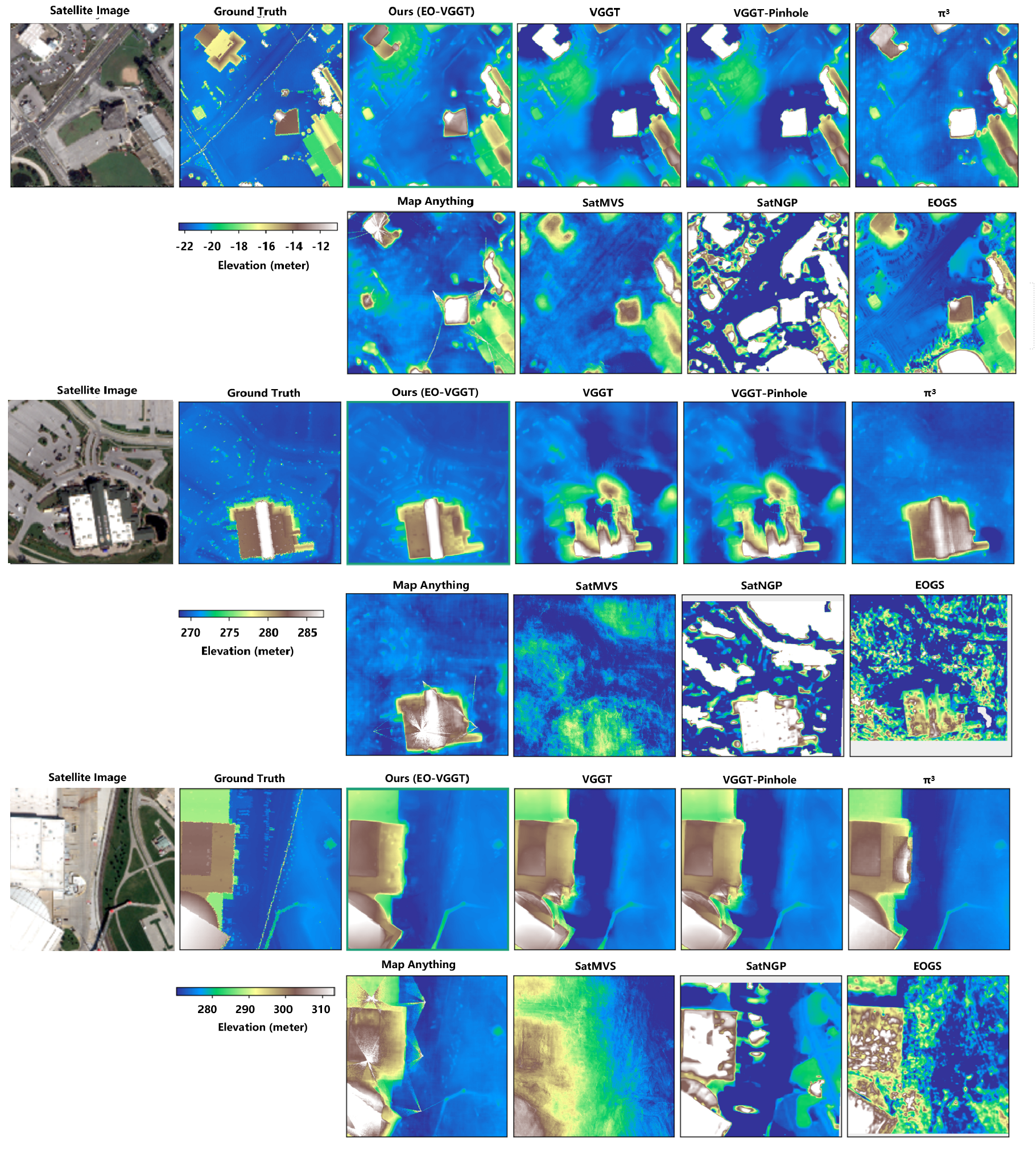}
  \caption{Qualitative visual comparison of the reconstructed macro-scale DSM structures across six representative geographical scenes spanning both the JAX and OMA domains. EO-VGGT successfully eliminates regional tilting and recovers sharp, well-defined building boundaries and flat terrain planes.}
  \label{fig:qual_all}
\end{figure*}
\endgroup

\subsection{Methodological Component Ablation}
To thoroughly disentangle the individual algorithmic contributions of the proposed components within the unified framework, we execute a rigorous $2 \times 3$ full factorial ablation matrix. As tabulated in Table~\ref{tab:component_ablation}, all configurations are evaluated at a fixed budget of $K=9$ and uniformly integrated via the spatial quadratic feathered blending merging pipeline to ensure strict algorithmic isolation. 


The experimental results lead to two key conclusions. First, the joint integration of orbital ray representation and embedding, implemented by the Sensor-Ray Encoder (SRE) and the Ray-Pointing-Aware Adapter (RPAA), is the primary factor in reducing structural errors. Under random view selection, activating the SRE-RPAA module reduces the All MAE from 2.526 meters to 1.861 meters, while under the GCCS strategy, it lowers the error from 2.437 meters to 1.751 meters. Second, without this geometric conditioning, deploying GCCS on the vanilla VGGT baseline yields a limited improvement, decreasing the All MAE from 2.526 meters to 2.437 meters. This confirms that view selection alone cannot bridge the fundamental sensor-model domain gap when applying computer vision models to satellite imagery. However, combining GCCS with the active SRE-RPAA framework achieves the lowest error profile across all metrics, yielding an All MAE of 1.751 meters and an All $P_{95}$ of 6.211 meters. These components function as complementary operations, where SRE and RPAA provide the core geographic adaptation by correcting the geometric sensor model, and GCCS establishes radiometrically stable multi-view sequences to safeguard the inference loop.

\subsection{Land-Cover Semantic Decomposition}
Table~\ref{tab:main_semantic} presents the macro-averaged MAE metrics across specific semantic land-cover strata, focusing on man-made buildings, natural ground terrain, and their spatial union. This analysis shows whether the performance enhancements are concentrated within specific land covers or shared across diverse topographies.

On the OMA test sets, EO-VGGT achieves consistent error reductions and lowers the building MAE to 1.175 meters. This performance outperforms the VGGT baseline value of 1.683 meters and the MapAnything baseline value of 1.511 meters, while improving upon the $\pi^3$ baseline value of 1.230 meters. For the natural ground regions that require long-range height consistency, our framework maintains an advantage, reducing the test ground MAE to 2.005 meters compared to 2.474 meters for the vanilla baseline. The validation split displays an elevated ground $P_{95}$ across all methodologies, which indicates the presence of low-texture terrain, dense vegetation, or potential local registration noise on the JAX split. We consider these factors as subjects for future work.

\subsection{Latent Space Cross-View Feature Consistency}
To evaluate the internal representation mechanism driving the surface reconstruction improvements, Table~\ref{tab:featcos_overall} quantifies the cross-view feature cosine consistency across physically grounded spatial correspondences. We report metrics under a natural spatial distribution using random sampling and under a class-balanced terrain profile using balanced sampling with 50 percent building pixels and 50 percent ground pixels to eliminate urban class-proportion biases across different target cities. 

As shown in Table~\ref{tab:featcos_overall}, under the balanced sampling protocol, the cross-view latent feature cosine similarity from our joint SRE-RPAA framework increases. Specifically, the feature similarity increases from 0.8837 to 0.9517 on the validation split and from 0.6146 to 0.8712 on the cross-city test split. Furthermore, the region-wise semantic decomposition in Table~\ref{tab:featcos_semantic} shows that these feature stabilization gains occur in both natural ground planes and man-made building boundaries. On the unseen OMA test split, the feature consistency for ground terrain and building structures exhibits relative improvements of 46.89 percent and 37.18 percent respectively. This indicates that our geometric conditioning framework prevents feature degradation when applying pretrained foundation vision models to satellite imagery.

\begin{figure}[!h]
  \centering
  \includegraphics[width=\linewidth]{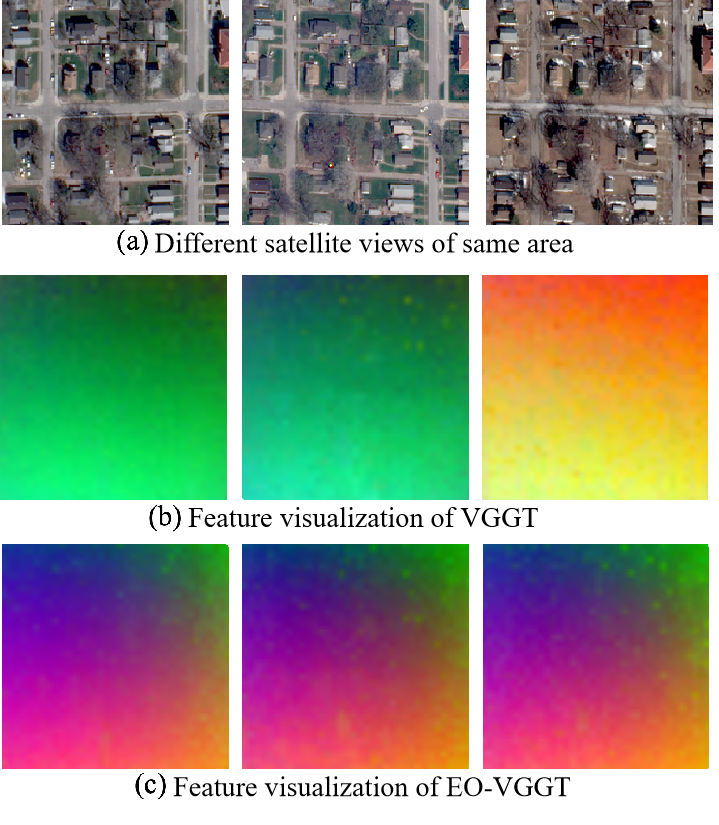}
  \caption{Visualization of last-layer latent feature structures mapping identical spatial features across different satellite view stacks using a shared Principal Component Analysis, or PCA, color basis for each row.}
  \label{fig:featcos_summary}
\end{figure}

Figure~\ref{fig:featcos_summary} shows the localized patch feature embeddings mapping identical spatial structures across different satellite view stacks, which are projected using a shared PCA color basis for each row. While the vanilla VGGT baseline exhibits feature color shifts and structural variations due to multi-temporal radiometric and illumination differences, EO-VGGT maintains cross-view semantic and color consistency. 

This outcome reveals the underlying mechanism of our method. By using the Sensor-Ray Encoder, or SRE, to map the spatially varying physical sensor lines of sight into high-dimensional geometric tokens, and then injecting them via the Ray-Pointing-Aware Adapter, or RPAA, into the frozen self-attention blocks, our framework aligns the latent representations of identical physical ground entities. Consequently, the internal foundation tokens become less sensitive to shadows, atmospheric variations, and multi-temporal appearance changes, which establishes a geometrically coherent reference frame for downstream dense 3D surface regression.

\subsection{View Budget Sensitivity and GCCS Robustness}
Table~\ref{tab:static_view_selection_k} presents a systematic evaluation of different view selection strategies across an operational budget scaling from $K = 3$ to $18$. The empirical results reveal a critical trade-off in satellite multi-view reconstruction between geometric consensus redundancy and multi-temporal noise propagation. For the random strategy, performance degrades severely and irregularly as $K$ increases, with its All MAE inflating from 1.830~m to 2.074~m. This baseline collapse contradicts traditional photogrammetric assumptions and highlights the vulnerability of frozen foundation backbones to uncurated latent feature noise in high-view regimes.

In the low-budget regime ($K \le 6$), the base\_angle\_diverse strategy yields a marginal edge in average errors (e.g., a global minimum All MAE of 1.748~m at $K=6$), benefiting from the strict exclusion of peripheral temporal variations. However, its accuracy plummets drastically once $K$ exceeds 9, with the All MAE skyrocketing to 1.949~m at $K=18$. This behavior underscores a severe scaling vulnerability when forced to ingest less optimal views. 

In contrast, the proposed GCCS strategy establishes an unassailable advantage in high-budget regimes ($K \ge 9$). It reaches its own optimal architectural configuration at $K=9$, balancing multi-view intersection consensus with sequence purity to deliver a tightly bounded All MAE of 1.751~m and an All $P_{95}$ of 6.211~m. More crucially, GCCS exhibits extraordinary scaling robustness; as $K$ scales from 9 to 18, its All MAE floats within a negligible margin of 0.011~m (1.751~m to 1.762~m), whereas competing baselines undergo catastrophic performance dissolution. These long-tail statistics validate that the joint radiometric and geometric constraints of GCCS successfully censor atmospheric and illumination anomalies, rendering it indispensable for stable, large-budget feed-forward 3D inference.



\subsection{Computational Overhead and Efficiency Profile}
Table~\ref{tab:runtime_p0} evaluates the computational efficiency and memory overhead of EO-VGGT across the complete benchmark of 66 AOIs. Because the foundation backbone is frozen, integrating the lightweight RPAA introduces low computational overhead. Specifically, activating the adapter increases the mean processing time per AOI from 24.4 to 25.1 seconds, which represents an increase of 0.7 seconds. Meanwhile, the peak GPU memory expands by 0.10 GiB, changing from 10.42 to 10.52 GiB. This low resource overhead, combined with the pre-computed GCCS view selection cache and tile blending, shows that EO-VGGT maintains high efficiency for large-scale processing.

\section{Discussion}

\subsection{Decoupling Geometric Mismatch from Radiometric Domain Gaps}
The difficulty in applying natural-image 3D foundation models to satellite multi-view reconstruction stems from both radiometric variations and differences in camera priors. Standard computer vision networks typically process multi-view features under the assumption of a central perspective projection. In contrast, satellite imagery follows the spatially varying trajectories of pushbroom sensors, which form a non-perspective projection manifold with multiple optical centers. 

Our framework addresses this mismatch by using the Sensor-Ray Encoder, or SRE, and the Ray-Pointing-Aware Adapter, or RPAA. This approach parameterizes Rational Function Model, or RFM, coordinates into high-dimensional tokens and introduces them as a learnable geometric condition. The relationship between feature-space consistency and geometric metrics supports this method. Improvements in latent cosine similarity indicate that our explicit ray conditioning stabilizes internal representations at projected correspondences. Concurrently, the digital surface model benchmarks show that this stability correlates with a reduction in aligned height residuals, indicating that geometric conditioning helps preserve structural fidelity.

\subsection{Relationship to Coordinate-Aware Geometric Deep Learning}
Explicitly embedding imaging geometry allows our framework to validate the consensus found in multi-view transformers that use ray-based coordinate encodings. These approaches generally suggest that deep networks perform better when camera geometry is provided rather than inferred entirely from data distributions. 

Our framework focuses on non-central satellite pushbroom geometry using a frozen, pretrained foundation backbone. Instead of retraining the core attention mechanisms for orbital geometries, the SRE-RPAA scheme parameterizes and injects normalized ray origins and directions for each patch. By using SRE to model satellite lines of sight and RPAA to route these cues through gated residual blocks, this architecture retains the spatially varying characteristics of the pushbroom sensor within localized spatial tiles while preventing the representation degradation common in full-parameter fine-tuning.

\subsection{Methodological Rationale for the Spatial Alignment Protocol}
The quantitative benchmarks in this evaluation measure surface reconstruction quality after a spatial alignment transformation. We use this protocol because our analysis focuses on whether a feed-forward foundation model can reconstruct local morphology, relative elevation differences, and structural continuity. This step matches established satellite reconstruction baselines that align the predicted terrain to reference data using a spatial transformation before computing residual metrics. 

Evaluating unaligned absolute elevation involves a different objective that combines structural reconstruction accuracy with horizontal geolocation offsets, vertical biases, sensor pose inaccuracies, and the availability of ground control points or bundle adjustment. Therefore, we separate absolute georegistration from our core architectural evaluation and treat absolute coordinate correction as a separate downstream task.

\subsection{Synergetic Effects of View Selection and Geometric Adaptation}
The unified configuration combines the SRE-RPAA geometric conditioning framework and the Geometry-Correlation Constrained Selection, or GCCS, strategy under a selected view budget, improving performance over both general foundation backbones and domain-specific pipelines. The ablation studies show that the joint SRE-RPAA adaptation provides the main reduction in structural errors, while the selection strategy offers additional refinement. Consequently, the selection strategy serves as a pre-inference filter to remove radiometrically or atmospherically inconsistent views before joint transformer processing rather than as a substitute for geometric adaptation. 

Our results show that surface completeness should be considered alongside residual error metrics in satellite mapping. Traditional stereoscopic pipelines can show low error metrics by omitting predictions in challenging regions, which leaves gaps over textureless or occluded zones. This pattern shows how local, handcrafted correspondence matching can be affected by multi-temporal atmospheric variations. By using the global context of the foundation transformer, EO-VGGT reduces these localized omissions to provide full-coverage topographies without omitting low-confidence regions from the error statistics.

\subsection{Information Saturation and Non-Monotonicity in View Aggregation}
Traditional satellite reconstruction pipelines often handle errors from individual stereo pairs during the volumetric or mesh fusion stages. In contrast, end-to-end multi-view foundation networks process input views together to update a shared latent representation. Under this joint inference approach, a small subset of radiometrically inconsistent or geometrically unstable views can affect the accuracy of the overall reconstruction. 

The GCCS strategy addresses this issue by using a normalized cross-correlation filter to reduce radiometric mismatches and then sampling the remaining candidates based on spatial intersection angles. This dual-constrained approach balances imaging consistency and geometric complementarity. 

The view-budget sensitivity analysis indicates that satellite multi-view foundation models do not always show a monotonic increase in accuracy when more views are added. Improving reconstruction accuracy with a larger number of views depends on managing corrupted or redundant views rather than simply increasing the input size. Because the angle-quantile selection mechanism is non-nested, the optimal view configurations for each budget level are generated independently.

\section{Limitations and Future Work}
Although EO-VGGT demonstrates robust performance, several areas can be extended in future work. On the JAX validation split, the ground region $P_{95}$ errors show an increase across all evaluated methods. Specifically, standard baselines exceed 23 meters while EO-VGGT registers 18.2 meters. This increase relates to the matching difficulties of optical sensors over low-texture or complex terrains in the JAX split, including water bodies, dense vegetation, and shadows from high-rise buildings, which are collectively classified as natural ground terrain. EO-VGGT reduces these geometric distortions by over 5 meters compared to the unconditioned baselines. On the cross-city OMA test split, the ground $P_{95}$ error decreases to 5.4 meters, which indicates that the higher errors are related to specific geographical characteristics rather than algorithmic instability. Future work will investigate integrating explicit smoothness priors or multi-spectral features to mitigate these low-texture matching challenges. We also plan to evaluate the framework across alternative foundation backbones and incorporate absolute georegistration blocks to improve absolute positioning accuracy.

\section{Conclusion}
In this paper, we present EO-VGGT, a framework that applies feed-forward neural multi-view reconstruction to satellite remote sensing. This approach addresses the error accumulation common in traditional photogrammetric pipelines and improves generalizability under geographical heterogeneity. By resolving the mismatch caused by implicit perspective camera assumptions and handling view heterogeneity from orbital constraints, the framework integrates computer vision foundation models with satellite imaging physics.

The joint Sensor-Ray Encoder, or SRE, and the lightweight Ray-Pointing-Aware Adapter, or RPAA, address the geometric mismatch by parameterizing and embedding Rational Function Model sensor rays into the frozen transformer backbone. This embedding helps recover an aligned latent reference frame with low parameter overhead. Concurrently, the Geometry-Correlation Constrained Selection, or GCCS, strategy reduces error propagation in the latent space by balancing geometric diversity with radiometric quality, which helps protect the scene-level geometric representations. Evaluations on the US3D benchmarks show accuracy improvements across both building and ground regions. These results confirm that combining pretrained structural priors with explicit physical geometry provides an effective framework for satellite mapping.

\section*{Acknowledgement}
This work was supported by the National Science Fund for Distinguished Young Scholars (62425102),  Hubei Province Strategic Talent Cultivation Project (2024DJA035), Open Fund of Hubei Luojia Laboratory(260100005) and LIESMARS Special Research Funding.

\begingroup
\small
\sloppy

\bibliographystyle{elsarticle-harv}
\bibliography{reference}  
\endgroup

\end{document}